\documentclass[pdflatex,sn-mathphys-num]{sn-jnl}

\usepackage{graphicx}%
\usepackage{multirow}%
\usepackage{amsmath,amssymb,amsfonts}%
\usepackage{amsthm}%
\usepackage{mathrsfs}%
\usepackage[title]{appendix}%
\usepackage{xcolor}%
\usepackage{textcomp}%
\usepackage{manyfoot}%
\usepackage{booktabs}%
\usepackage{algorithm}%
\usepackage{algorithmicx}%
\usepackage{algpseudocode}%
\usepackage{listings}%
\usepackage[colorinlistoftodos]{todonotes}
\usepackage{setspace}
\usepackage{geometry}
\usepackage[most]{tcolorbox}

\tcbset {
  base/.style={
    arc=0mm, 
    bottomtitle=0.5mm,
    boxrule=0mm,
    colbacktitle=black!10!white, 
    coltitle=black, 
    fonttitle=\bfseries, 
    left=2.5mm,
    leftrule=1mm,
    right=3.5mm,
    title={#1},
    toptitle=0.75mm, 
  }
}

\newtcolorbox{mytextbox}[1][]{%
  sharp corners,
  enhanced,
  colback=white,
  height=2.3cm,
  attach title to upper,
  #1
}

\newtcolorbox{mytextbox2}[1][]{%
  sharp corners,
  enhanced,
  colback=white,
  height=5.0cm,
  attach title to upper,
  #1
}

\definecolor{brandblue}{rgb}{0.34, 0.7, 1}
\newtcolorbox{mainbox}[1]{
  colframe=brandblue, 
  base={#1}
}
\renewcommand\fbox{\fcolorbox{red}{white}}
\newtcolorbox{subbox}[1]{
  colframe=black!30!white,
  base={#1}
}

\theoremstyle{thmstyleone}%
\theoremstyle{thmstyletwo}%

\theoremstyle{thmstylethree}%

\begin{document}

\title[Article Title]{Revisiting Gene Ontology Knowledge Discovery with Hierarchical Feature Selection and Virtual Study Group of AI Agents}


\author*[1]{\fnm{Cen} \sur{Wan}}\email{Cen.Wan@bbk.ac.uk}

\author[2]{\fnm{Alex A.} \sur{Freitas}}\email{A.A.Freitas@kent.ac.uk}


\affil*[1]{\orgdiv{School of Computing and Mathematical Sciences}, \orgname{Birkbeck, University of London}, \orgaddress{\city{London}, \country{United Kingdom}}}

\affil[2]{\orgdiv{School of Computing}, \orgname{University of Kent}, \orgaddress{\city{Canterbury}, \country{United Kingdom}}}



\abstract{Large language models have achieved great success in multiple challenging tasks, and their capacity can be further boosted by the emerging agentic AI techniques. This new computing paradigm has already started revolutionising the traditional scientific discovery pipelines. In this work, we propose a novel agentic AI-based knowledge discovery-oriented virtual study group that aims to extract meaningful ageing-related biological knowledge considering highly ageing-related Gene Ontology terms that are selected by hierarchical feature selection methods. We investigate the performance of the proposed agentic AI framework by considering four different model organisms’ ageing-related Gene Ontology terms and validate the biological findings by reviewing existing research articles. It is found that the majority of the AI agent-generated scientific claims can be supported by existing literatures and the proposed internal mechanisms of the virtual study group also play an important role in the designed agentic AI-based knowledge discovery framework.}



\maketitle

\section{Introduction}
The success of large language models (LLMs) and the emerging agentic AI techniques have already started revolutionalising the traditional scientific discovery paradigm. One of the most well-known examples is the agentic AI-based virtual lab \cite{Swanson}, where real nanobody binders were successfully designed by AI agents autonomously. This achievement not only validated the feasibility of agentic AI-based scientistic research pipelines, but also boosted the research in exploring agentic AI’s possibilities in other research areas. 

As one of the core topics in scientific research, knowledge extraction and discovery play a fundamental role yet remain a major challenge in almost all research subjects like ageing biology, which aims to explore extremely complex theories involving enormously unrevealed biological mechanisms. Thanks to the efforts made by the community, new progress has been recently made by AI researchers who predict novel knowledge about ageing. For example, Alsaggaf, et al (2024) \cite{AgeingPrediction1} developed novel contrastive learning algorithms to successfully predict novel ageing-related genes using protein-protein interaction networks, whilst Paz-Ruza, et al (2024) \cite{AgeingPrediction2} also predicted novel dietary restriction-related genes by using positive-unlabelled learning algorithms and human gene-pathway associations information. However, the limited knowledge about ageing processes remains a bottleneck of any predictive models that need to exploit sufficient information to obtain reliable performance. 

Large language models have recently demonstrated strong capacity in handling complicated tasks (e.g. automatic computer programming code generation \cite{CodeGenerationLLM2}), but their own shortcomings due to the well-known hallucination issues and lack of model interpretability still lead to concerns about its application in knowledge discovery tasks. The capacity of LLMs can be further enhanced by integrating the agentic AI approach to handle more challenging tasks (e.g. strategic planning for auctions \cite{PlanningLLM}). Hence, LLM-based agentic AI frameworks that allow multiple LLMs to mimic human interactions would be a feasible research direction aiming to alleviate the problems of a single LLM-based knowledge discovery method. In this work, we propose a novel LLM-based agentic AI framework that simulates a type of human study group mechanism for extracting meaningful ageing-related biological knowledge according to informative Gene Ontology terms selected by hierarchical feature selection methods.

\begin{figure*}[!t]
    \centering
    \includegraphics[width=16.5cm]{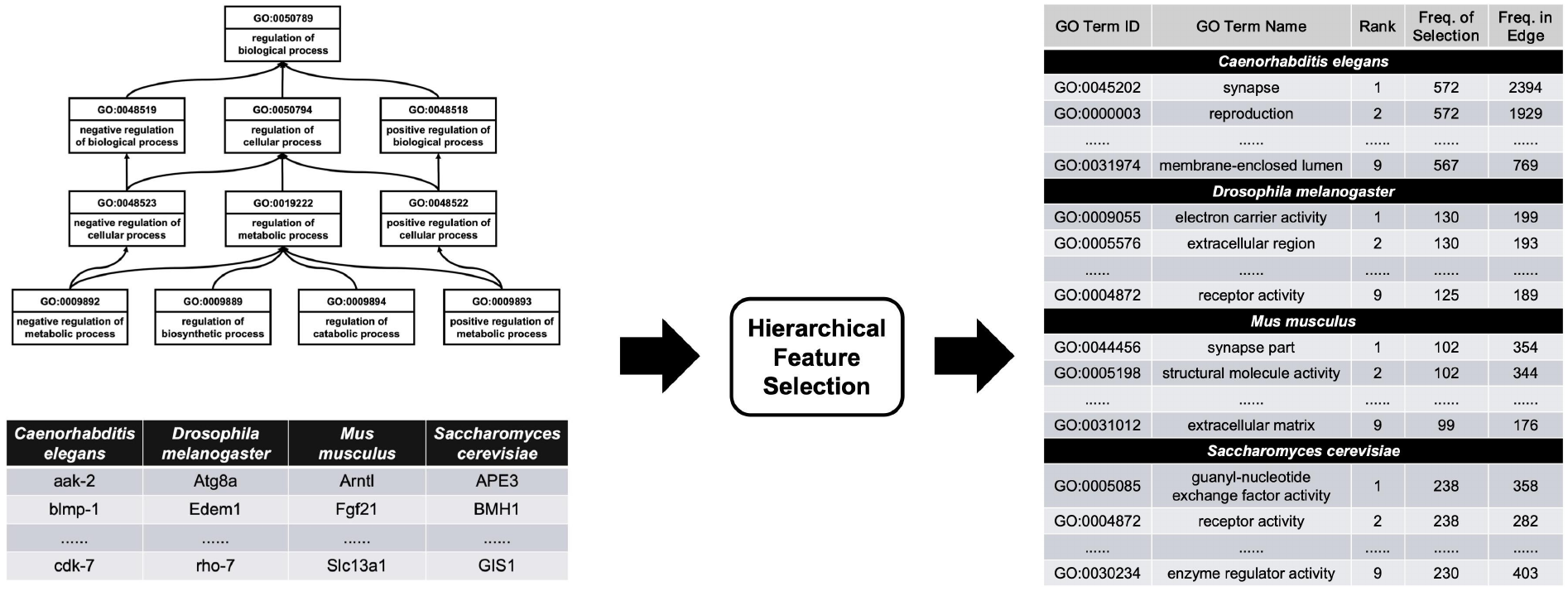}
    \vspace{0.5cm}
    \caption{An example of hierarchical feature selection-based Gene Ontology knowledge discovery pipeline.}
     \vspace{0.3cm}
    \label{fig1}
\end{figure*}
\vspace{0.3cm}
\section{Background}
\subsection{Gene Ontology Database}
Gene Ontology database \cite{GeneOntology} provides a type of universal vocabulary, i.e. Gene Ontology terms, to describe functions of genes and gene products. All GO terms are organised as a type of hierarchical structure, where GO terms that are located in higher positions denote more generic definitions and those are located in lower positions denote more specific definitions. As shown in the example GO hierarchy in Figure 1, GO:0050789 that denotes regulation of biological process is located in the highest position, while other GO terms, e.g. GO:0009894 (regulation of catabolic process), are located in lower positions, denoting more specific definitions of GO:0050789. Therefore, with considering genes and their corresponding Gene Ontology annotations, we are able to use a set of GO terms to describe those genes' biological processes, molecular functions and cellular components. In addition, due to the hierarchical relationships between GO terms, when conducting biological knowledge interpretation, those GO terms bearing more specific definitions are more informative than the ones bearing more generic definitions. However, note that those GO terms bearing more specific definitions are usually independent to other GO terms that also denote specific definitions, leading to a challenge to discover associations between multiple GO terms bearing specific biological definitions. For example, as shown in Figure 1, GO:0009892 (negative regulation of metabolic process) and GO:0009889 (regulation of biosynthetic process) that denote two different specific biological processes have no interconnections according to the pre-defined hierarchy. Therefore, it will be difficult to find the interaction of the biological definitions denoted by those two GO terms, when some genes are simultaneously annotated by both of them.

\subsection{Hierarchical Feature Selection}
Hierarchical feature selection \cite{Wan2013,Petar2014,Wan2015,WanBook,Pablo2018,Pablo2020,Dehui2022,Wan2024,Jianyun2025} is a type of dimensionality reduction methods that aim to remove redundant and irrelevant features with considering pre-defined hierarchical dependencies. Wan \& Freitas firstly proposed a family of hierarchical feature selection methods including HIP (Select Hierarchical InformationPreserving features) \cite{Wan2015}, MR (Select Most Relevant features) \cite{Wan2015} and HIP-MR (the hybrid of HIP and MR methods) \cite{Wan2013} that were used to deal with the Gene Ontology database, where GO terms were considered as features. With the help of the proposed methods, a set of the most informative GO terms associated with ageing-related genes \cite{HAGR} were selected and ranked as a table, which provided a knowledge source for biological interpretations. For example, as shown in Figure 1, with considering four different sets of genes that have been found as being relevant to ageing processes, hierarchical feature selection methods like HIP can derive a table showing the most informative GO terms for different model organisms. In terms of \textit{Caenorhabditis elegans}, the most informative GO term is GO:0045202 (synapse), followed by the second top-ranked GO term GO:0000003 (reproduction). Note that, both GO:0045202 and GO:0000003 have the same frequencies of selection. Therefore, I used frequency in edges as a tie-breaking criterion, i.e. tree-augmented naive Bayes \cite{TAN} was used to learn a tree for each instance by using the corresponding selected GO terms. Analogously, in terms of \textit{Drosophila melanogaster}, the top-ranked GO terms are GO:0009055 (electron carrier activity) and GO:0005576 (extracellular region). In this work, I use the table reported in Wan \& Freitas (2018) \cite{Wan2018} that was generated by the HIP method as an example to evaluate the proposed AI agent-based virtual study group.

\subsection{Large Language Models for Knowledge Extraction}
Large Language Models are a type of AI intelligent system trained by using massive and complex data with advanced language model architectures and techniques. One of the most impressive properties of LLMs is the so-called knowledge emergence phenomenon \cite{EmergentAbility,SurveyLLM} that allows LLMs to convert multi-dimensional and multi-scale data into their capacity to conduct advanced activities, e.g. programming code generation \cite{CodeGenerationSurvey,CodeGenerationLLM,CodeGenerationLLM2}, logical reasoning inference \cite{ReasoningSurvey,ReasoningLLM2,ReasoningLLM}, and strategic planning \cite{PlanningSurvey,PlanningLLM,PlanningLLM2}. Due to this unique property, LLMs have demonstrated great potential in knowledge extraction tasks. For example, Dagdelen, et al. (2024) \cite{Dagdelen} successfully adopted LLMs to extract a type of structured information consisting of entities and their relationships from raw text. Analogously, Balasubramanian, et al. (2025) \cite{Balasubramanian} adopted LLMs to successfully extract knowledge that was presented as a set of features from pathology reports. Builtjes, et al. (2025) \cite{Builtjes} also proposed a LLM-based framework for extracting information from clinical texts. However, note that, all those above works are quite different with the Gene Ontology terms-based knowledge interpretation tasks we discuss in this work. Also, due to the well-known shortcomings of existing LLMs, e.g. the hallucination issue, it is unclear whether LLMs are able to derive reliable biological knowledge that usually involves complex theories.

\subsection{Agentic AI for Scientific Research}
Agentic AI is a type of intelligent system that evolves from the conventional multi-agent systems \cite{MultiagentBook} can autonomously conduct activities to achieve pre-defined goals. Compared with the conventional multi-agent systems, agentic AI systems are more powerful, due to the superbo capacities of large language models that are used as cores. An agentic AI system usually consists of multiple AI agents, which form a virtual team where each agent plays a role allowing them to cooperate with each other by passing information and calling different tools until the pre-defined goal is achieved. Recently, multiple agentic AI systems have been developed for scientific discovery tasks. For example, Ruan, et al. (2024) \cite{LLMRDF} developed a LLM-agent-based end-to-end chemical synthesis development platform that successfully conducts multiple chemical research steps including literature search, reaction kinetics study, and product purification. More recently, Swanson, et al. (2025) \cite{Swanson} proposed an AI agent-based virtual lab to successfully design nanobody binders. Within the virtual lab, multiple agents simulate human-like scientific discovery activities including group discussion, computational tools implementation and computational experiment results analysis. Analogously, Schmidgall, et al. (2025) \cite{Schmidgall} also developed an agent-based virtual lab that autonomously executes a completed research pipeline. Given a research idea, the proposed virtual lab can autonomously conduct literature review, research plan generation, programming code generation and results analysis, leading to a great decrease in research expense. It has also been found that the agentic AI paradigm can alleviate the hallucination issue. For example, Gosmar and Dahl (2025) \cite{Gosmar2025} proposed a multi-layer framework, where agents in higher layers aim to remove hallucination information in the statements made by agents in lower layer. In this work, we also propose a multi-layer agent framework, which aims to exploit multiple agents to process LLM-generated biological knowledge by using a type of critical reviewing mechanisms to extract meaningful ageing-related knowledge considering certain biological processes and molecular functions.

\vspace{0.3cm}
\section{Problem Definition}
In this paper, we focus on interpreting ageing-related biological knowledge from a set of informative Gene Ontology terms that are selected by hierarchical feature selection methods. Given a list of selected Gene Ontology terms according to gene annotations for four different model organisms, an agentic AI system is required to generate a report describing biological findings about each individual model organism and common patterns across those organisms.  \newline

\noindent \textit{More specifically, we define the tasks as shown below.}\newline

\begin{mytextbox}
\begin{itemize}
    \item Evaluate the correctness of GO term-based biological knowledge interpretations generated by LLMs.\newline
    \item Evaluate the usefulness of the proposed agentic AI-based virtual study group mechanisms for biological knowledge discovery.
\end{itemize}
\end{mytextbox}
\vspace{0.3cm}
\noindent \textit{The input GO terms and the tasks assigned to different AI agents are shown below. In this work, we consider the GO terms selection table reported in our previous work \cite{Wan2018}. In terms of the GO terms used in this work, we select the ones with high selection frequency and bear more generic biological definitions which would be feasible to be validated by existing literatures.}\newline
\begin{itemize}
    \item \textbf{Worm:} 
              \begin{itemize}
    		   \item	\textbf{Input GO terms:} \{GO:0000003 (reproduction), GO:0016209 (antioxidant activity)\}\vspace{0.3cm}
		   \item      \textbf{Tasks:} Investigate the associations between reproduction, antioxidant activity and ageing processes in \textit{Caenorhabditis elegans}.\vspace{0.3cm}
		   \item      \textbf{Remarks:} both two terms are top-ranked GO terms that are selected by the HIP hierarchical feature selection method for all 572 ageing-related worm genes in the given dataset. The two GO terms have no connections in the GO hierarchy and belong to different domains, i.e. the former belongs to the Biological Process domain, and the latter belongs to the Molecular Function domain.\newline
    	      \end{itemize}
    
        \item \textbf{Fruit fly:} 
              \begin{itemize}
    		   \item	\textbf{Input GO terms:} \{GO:0009055 (electron carrier activity), GO:0005198 (structural molecule activity)\}\vspace{0.3cm}
		   \item      \textbf{Tasks:} Investigate the associations between electron carrier activity, structural molecule activity and ageing processes in \textit{Drosophila melanogaster}.\vspace{0.3cm}
		   \item      \textbf{Remarks:} both two terms are top-ranked GO terms that are selected by the HIP hierarchical feature selection method for ageing-related fruit fly genes in the given dataset. The two GO terms have no connections in the GO hierarchy, though both belong to the Molecular Function domain.\newline
	 \end{itemize}
       
       \item \textbf{Mouse:} 
              \begin{itemize}
    		   \item	\textbf{Input GO terms:} \{GO:0005198 (structural molecule activity), GO:0016209 (antioxidant activity)\}\vspace{0.3cm}
		   \item      \textbf{Tasks:} Investigate the associations between structural molecule activity, antioxidant activity and ageing processes in \textit{Mus musculus}.\vspace{0.3cm}
		   \item      \textbf{Remarks:} both two terms are top-ranked GO terms that are selected by the HIP hierarchical feature selection method for ageing-related mouse genes in the given dataset. The two GO terms have no connections in the GO hierarchy, though both belong to the Molecular Function domain.\newline
    	      \end{itemize}
       \item \textbf{Yeast:} 
              \begin{itemize}
    		   \item	\textbf{Input GO terms:} \{GO:0004872 (receptor activity), GO:0022414 (reproductive process)\}\vspace{0.3cm}
		   \item      \textbf{Tasks:} Investigate the associations between receptor activity, reproductive process and ageing processes in \textit{Saccharomyces cerevisiae}.\vspace{0.3cm}
		   \item      \textbf{Remarks:} both two terms are top-ranked GO terms that are selected by the HIP hierarchical feature selection method for ageing-related yeast genes in the given dataset. The two GO terms have no connections in the GO hierarchy and belong to different domains, i.e. the former belongs to the Molecular Function domain, and the latter belongs to the Biological Process domain.
    	      \end{itemize}
\end{itemize}

\newpage
\section{Proposed Methods}
We propose an AI agents-based Virtual Study Group (VSG) framework, where the biological interpretations of the Gene Ontology terms selected by hierarchical feature selection methods can be extracted automatically. The proposed VSG framework is composed by several agents, which play different roles in a hierarchically organised virtual organization. In order to extract a general summary with considering all GO terms' biological definitions, we propose a bottom-up structure that is used to coordinate the actions of multiple agents. As shown in Figure 2, in the first layer, multiple virtual junior researchers that mimic early year PhD students are in charge of investigating the associations between different GO terms and ageing processes. As we have four model organisms in this case, VSG uses four virtual junior researchers which have certain academic background on the corresponding model organims. For example, one virtual early-year PhD student (i.e. Virtual Junior Researcher A) whose academic background is mainly about worm will be assigned to conduct early investigations on the associations between certain GO terms and worm-related ageing processes. In addition, for each of the virtual junior researcher, we assign another virtual junior researcher (i.e. Virtual Junior Researcher B) whose duty is to criticise the findings discovered by Virtual Junior Researcher A. The second layer consists of four virtual senior researchers mimicking post-doctoral research associates, who are in charge of summarising the findings and the corresponding critics made by two virtual junior researchers (Virtual Junior Researcher A and Virtual Junior Researcher B) for one model organism. After obtaining the reports made by four virtual senior researchers, a virtual principal investigator that mimic an ageing professor will provide a final report with considering the findings about all four model organisms.

We implement the proposed VSG system using CrewAI \cite{CrewAI} and four different open-source large language models that are held by Ollama \cite{Ollama}. As shown in Figure 3, we use deepseek-r1 \cite{DeepSeek} and qwen3-vl \cite{Qwen3VL} as the cores for Virtual Junior Researcher A and Virtual Junior Researcher B, respectively. gpt-oss  \cite{gptoss} and glm-4.7-flash \cite{GLM47Flash} are used as the cores for Virtual Senior Researchers and Virtual Principal Investigator, respectively.

\begin{figure*}[!t]
    \centering
    \includegraphics[width=16cm]{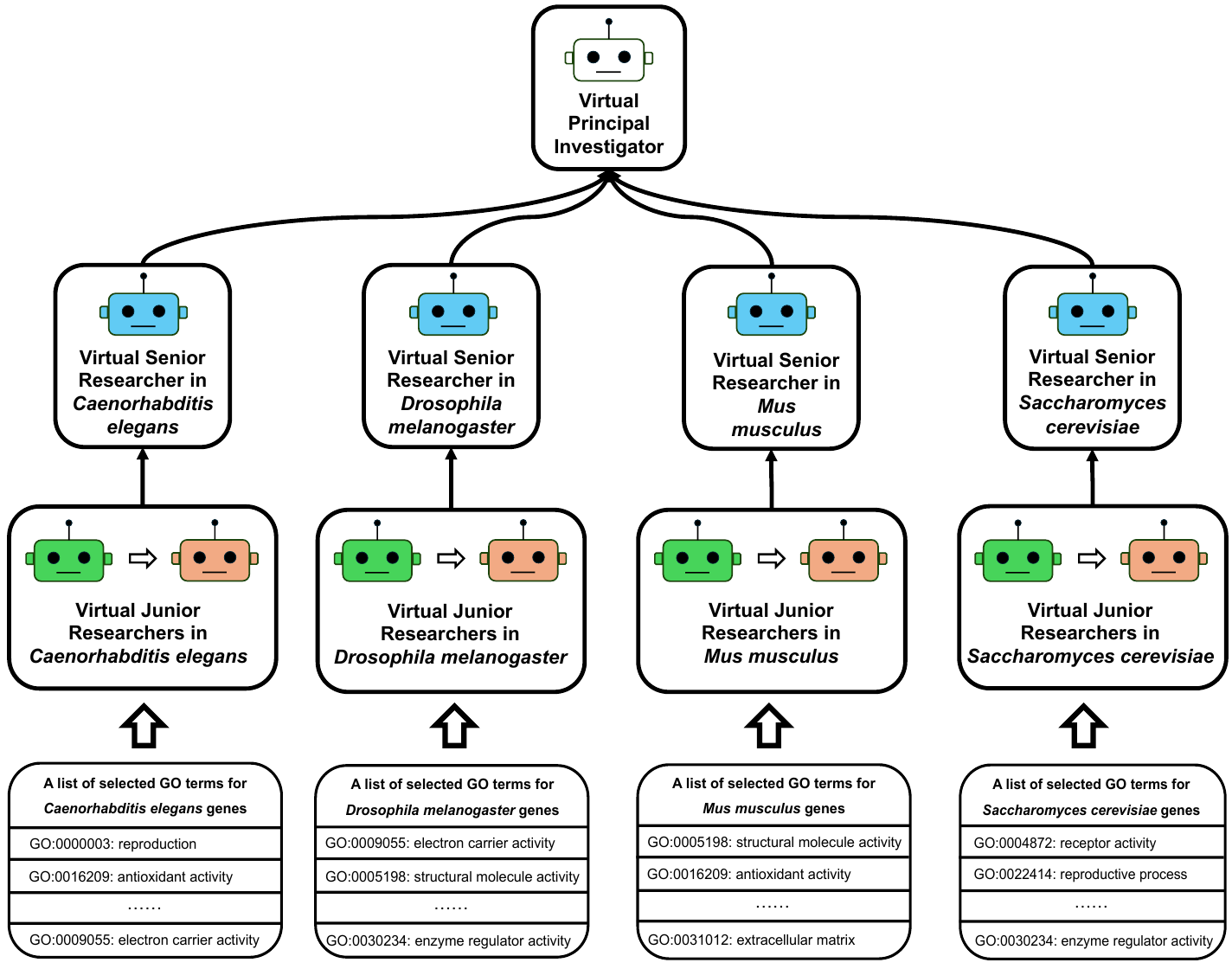}
    \vspace{0.3cm}
    \caption{The hierarchical structure of the proposed AI agents-based virtual study group framework.}
    \label{fig1}
\end{figure*}

\begin{figure*}[!t]
    \centering
    \includegraphics[width=16cm]{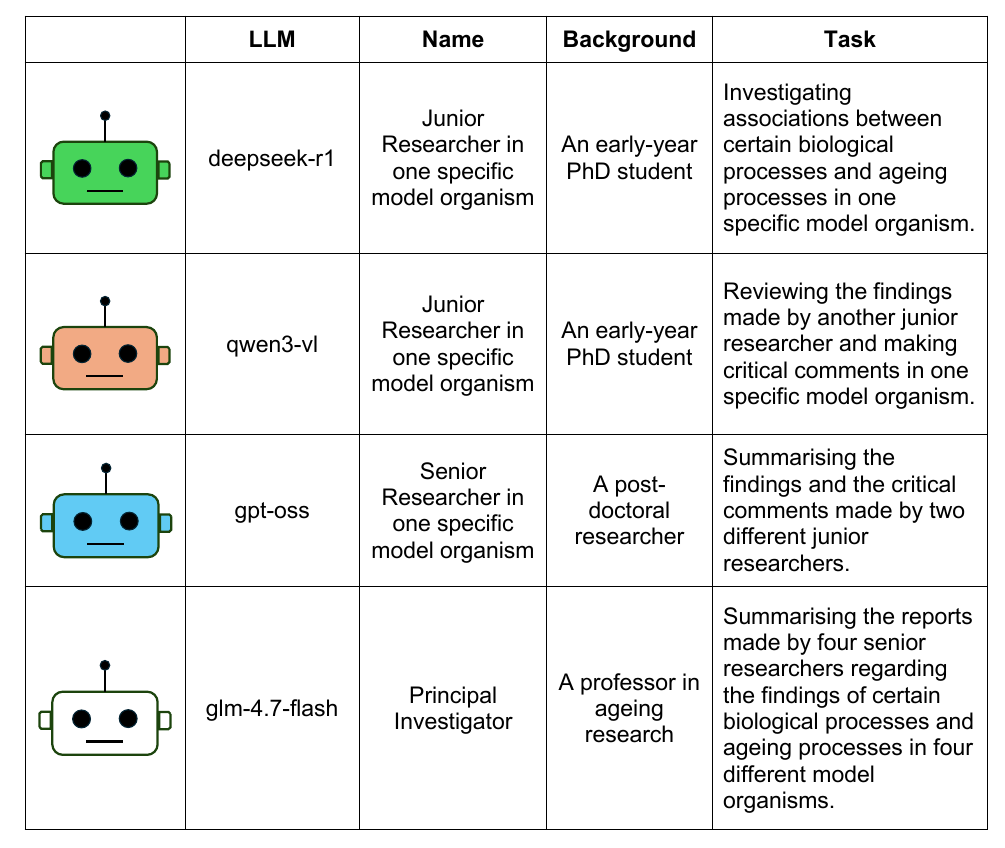}
    \vspace{0.3cm}
    \caption{Further information regarding different LLM-based agents.}
    \label{fig1}
\end{figure*}

\newpage\newpage
\section{Experimental Results}
\subsection{The majority of the claims made by virtual junior researchers are supported by known biological knowledge.}
We report the original outputs made by different AI agents and highlight any claims that can be validated by existing biological knowledge through conducting literature reviews on existing research articles published in decent academic journals. The phrases that are highlighted in green backgrounds denote the scientific claims that are supported by at least one existing research article, whilst the phrases in yellow backgrounds denote the scientific claims that are not supported by any existing research articles. 

In general, the generated reports by AI agents include both objective claims and subjective opinions. In terms of Virtual Junior Worm Researcher A and Virtual Junior Worm Researcher B, the majority of the scientific and objective claims are correct and can be validated by existing research articles. For example, Virtual Junior Worm Researcher A successfully points out that superoxide dismutase (SOD) and catalase (CAT) play a crucial role in antioxidant defense systems that are tightly associated with ageing. Virtual Junior Worm Researcher B also successfully further mentions the role of sod-3 in hypodermis in terms of age-related regulations. Analogously, Virtual Junior Fruit Fly Researcher A mentions the associations between electron carriers and mitochondrial function, and Virtual Junior Fruit Fly Researcher B further explains the role of UPR$^{mt}$ in ageing processes. In terms of Virtual Junior Mouse Researcher A, it successfully points out the associations between laminopathies and progeria, whilst Virtual Junior Mouse Researcher B further mentions the role of nuclear instability in mouse's ageing processes. Finally, in terms of Virtual Junior Yeast Researcher A, the well-known associations between mating pheromones and yeast ageing processes are successfully revealed, whilst Virtual Junior Yeast Researcher B further points out the role of quiescent cell metabolism, which is highly associated with yeast ageing processes. However, note that Virtual Junior Yeast Researcher B claims that `Hog1 deletion mutants maintain near-normal chronological lifespan in standard conditions'. This claim is not supported and an opposite conclusion was made by one existing research article.

\subsection{Contrary views between virtual junior researchers are successfully observed and senior researchers made fair points.}
In terms of the subjective opinions, the Virtual Junior Worm Researcher A, Virtual Junior Fruit Fly Researcher A, Virtual Junior Mouse Researcher A and Virtual Junior Yeast Researcher A all use a natural tone of certainty to summarise the biological knowledge findings by providing the corresponding objective evidence. On the contrary, the Virtual Junior Worm Researcher B, Virtual Junior Fruit Fly Researcher B, Virtual Junior Mouse Researcher B and Virtual Junior Yeast Researcher B all criticise the opinions made by their counterparts and also show corresponding objective evidence. For example, some commonly used terms are oversimplifies, overstates, and overgeneralizes, suggesting a type of critical context. 

In terms of those senior researchers for the four model organisms, all of them tend to agree with the criticisms made by the corresponding virtual junior researchers B by rephrasing the content provided by virtual junior researchers A and B. In addition, those virtual senior researchers all suggest that more research needs to be conducted to further investigate the ageing mechanisms for different model organisms.

The comments made by the virtual ageing professor are also interesting. To begin with, it successfully points out that perhaps the more valuable direction would be related to the generalisability of the worm-based knowledge in terms of extending to mammals due to the complexity of the latter. Moreover, it acknowledges and reiterates the importance of Reactive Oxygen Species and their related biological pathways in ageing processes, but also points out the challenges of future research in this direction. Furthermore, it mentions that the discussions about the associations between reproduction and ageing need to be further expanded from evolutionary perspectives.

\begin{tcolorbox}[colback=green!5,colframe=green!40!black,fontupper=\small,title={
\vspace{0.3cm}
\begin{itemize}
  \item \small Agent: Virtual Junior Worm Researcher A
  \item Task: Investigate the associations between reproduction, antioxidant activity and ageing processes in \textit{Caenorhabditis elegans}.
\end{itemize}
\vspace{0.3cm}
},code={\doublespacing}]
\textbf{AI Agent Outputs:}\newline
Reproduction in *\textit{C. elegans}* is a critical determinant of lifespan, largely occurring during a specific reproductive period. This reproductive activity, particularly within the germline, \colorbox{green}{is associated with increased production of reactive oxygen species (ROS) \fbox{supported by \cite{CElegans1}}}, \colorbox{green}{a form of cellular damage linked to aging. Consequently, the efficiency of antioxidant defense systems,} \colorbox{green}{which include enzymes like superoxide dismutase (SOD) and catalase (CAT), becomes crucial.} \newline \colorbox{green}{\fbox{supported by \cite{CElegans2}}}  \colorbox{green}{Reduced antioxidant activity or impaired antioxidant pathways, such as those}  \colorbox{green}{regulated by the DAF-16/FOXO transcription factor downstream of the insulin/IGF-1 signaling (IIS)} \colorbox{green}{pathway, can accelerate aging and reduce lifespan. \fbox{supported by \cite{CElegans3,CElegans4}}} Conversely, enhanced antioxidant capacity can promote longevity. The relationship is complex, involving trade-offs; reproductive investment might influence resource allocation to maintenance pathways, including antioxidant defenses. \colorbox{green}{Mutants with defects in reproduction or longevity pathways (e.g., *clk-1* \fbox{supported by \cite{CElegans5}},} \colorbox{green}{*germline-less* \fbox{supported by \cite{CElegans6}}) often exhibit altered oxidative stress levels}, highlighting the intertwined nature of reproduction, antioxidant status, and the aging process in worms.    \newline
\noindent\rule{15.3cm}{0.4pt}         \newline
\fbox{\textbf{* Authors' comments:} the generated report can be validated by existing research articles.} 
\end{tcolorbox}

\begin{tcolorbox}[colback=orange!5,colframe=orange!40!black,fontupper=\small,title={
\vspace{0.3cm}
\begin{itemize}
  \item \small Agent: Virtual Junior Worm Researcher B
  \item Task: Provide critical comments against the report made by Virtual Junior Worm Researcher A.
\end{itemize}
\vspace{0.3cm}
},code={\doublespacing}]
\textbf{AI Agent Outputs:}\newline
The received information oversimplifies the relationship between reproduction, oxidative stress, and aging in *\textit{C. elegans}* by framing reactive oxygen species (ROS) as universally detrimental without acknowledging their dual role as signaling molecules that regulate stress responses and longevity pathways. Specifically, it erroneously implies that germline-derived ROS universally accelerate aging, \colorbox{green}{while empirical evidence from *germline-less* mutants (e.g., *glp-1* RNAi) demonstrates that} \colorbox{green}{*reducing* germline activity extends lifespan *despite* elevated systemic ROS \fbox{supported by \cite{CElegans7,CElegans6}}}, contradicting the assertion that reproductive investment directly increases oxidative damage. Furthermore, the text conflates correlation with causation by attributing altered oxidative stress in mutants like *clk-1* solely to reproductive defects, whereas \colorbox{green}{*clk-1* primarily affects mitochondrial function and metabolism independently of germline activity,} \colorbox{green}{and its lifespan extension is linked to reduced ROS in somatic tissues -- not germline-specific} \colorbox{green}{trade-offs \fbox{supported by \cite{CElegans8,CElegans9}}}. The discussion of antioxidant defense efficiency overlooks tissue-specific dynamics, such as the critical role of \colorbox{green}{*sod-3* in hypodermis \fbox{supperted by \cite{CElegans10}}} or \colorbox{yellow}{*cat-1* in neurons \fbox{support not found}}, which are not equally impacted by reproductive status. Additionally, the claim that DAF-16/FOXO activation consistently suppresses ROS ignores context-dependent mechanisms, as \colorbox{green}{DAF-16 can induce *sod-3* in certain conditions but not others \fbox{supported by \cite{CElegans11}}}, and \colorbox{green}{germline removal activates DAF-16 independently of IIS in *germline-less* adults \fbox{supported by \cite{CElegans12}}}. Crucially, the text neglects that \colorbox{green}{ROS production in the germline serves essential functions (e.g., meiotic} \colorbox{green}{recombination) \fbox{supported by \cite{CElegans21}}}, and the described ``trade-off" model fails to account for compensatory pathways, such as the role of \colorbox{green}{mitochondrial retrograde signaling \fbox{supported by \cite{CElegans16}}} or \colorbox{green}{dietary interventions \fbox{supported by \cite{CElegans17}}} (e.g., \colorbox{green}{rapamycin \fbox{supported by \cite{CElegans13}}}) that modulate both reproduction and oxidative stress without direct germline involvement. This oversimplification risks misrepresenting the complexity of aging mechanisms and underestimates the need for nuanced, tissue-resolved analysis in future studies.  \newline
\noindent\rule{15.3cm}{0.4pt}         \newline
\fbox{\textbf{* Authors' comments:} the criticisms are sensible and the majority of the scientific claims can be validated} 
\fbox{by existing research articles.} 
\end{tcolorbox}

\begin{tcolorbox}[colback=blue!5,colframe=blue!40!black,fontupper=\small,title={
\vspace{0.3cm}
\begin{itemize}
\item \small Agent: Virtual Senior Worm Researcher 
\item Task: Provide critical comments against the reports made by Virtual Junior Worm Researcher A and Virtual Junior Worm Researcher B.
\end{itemize}
\vspace{0.3cm}
},code={\doublespacing}]
\textbf{AI Agent Outputs:}\newline
The received analysis overstates the deleterious nature of ROS by treating them as uniformly harmful, ignoring the well‑documented signaling roles that ROS play in activating stress‑responsive and longevity pathways in *\textit{C. elegans}*; for instance, \colorbox{green}{germline‑less *glp‑1* mutants extend lifespan despite elevated systemic ROS \fbox{supported by \cite{CElegans7,CElegans6}}}, contradicting the claim that reproductive investment inevitably raises oxidative damage, and suggesting instead a complex reallocation of resources that can enhance systemic antioxidant defenses; the discussion also mistakenly equates altered oxidative stress in *clk‑1* mutants with reproductive defects, whereas \colorbox{green}{*clk‑1* primarily} \colorbox{green}{perturbs mitochondrial biogenesis and metabolism, and its lifespan extension is linked to reduced somatic ROS} \colorbox{green}{rather than germline trade‑offs \fbox{supported by \cite{CElegans8,CElegans9}}}; tissue‑specific nuances are overlooked, such as the pivotal role of \colorbox{green}{*sod‑3* in the hypodermis \fbox{supperted by \cite{CElegans10}}} and \colorbox{yellow}{*cat‑1* in neurons \fbox{supperted not found}}, which are differentially regulated by germline status and cannot be generalized across the organism; moreover, the assertion that DAF‑16/FOXO activation uniformly suppresses ROS is inaccurate, as \colorbox{green}{DAF‑16 can induce *sod‑3* only under certain conditions \fbox{supported by \cite{CElegans11}}} and \colorbox{green}{germline removal activates DAF‑16 via IIS‑independent mechanisms \fbox{supported by \cite{CElegans12}}}, indicating context‑dependent pathways; the critique also neglects the \colorbox{green}{essential germline functions of ROS in meiotic} \colorbox{green}{recombination \fbox{supported by \cite{CElegans21}} and chromosomal segregation \fbox{supported by \cite{CElegans22}}}, and it fails to account for compensatory mechanisms such as \colorbox{green}{mitochondrial retrograde signaling \fbox{supported by \cite{CElegans16}}} or \colorbox{green}{dietary interventions \fbox{supported by \cite{CElegans19}}} (e.g., \colorbox{green}{rapamycin \fbox{supported by \cite{CElegans20}}}) that modulate both reproductive output and oxidative stress independently of germline activity; overall, the simplified model presented risks misguiding future research by underappreciating the multi‑layered, tissue‑resolved interactions between reproduction, oxidative stress, and longevity in *\textit{C. elegans}* and highlights the need for nuanced, experimentally dissected analyses rather than broad generalizations.      \newline
\noindent\rule{15.3cm}{0.4pt}         \newline
\fbox{\textbf{* Authors' comments:} the report mainly agrees with the criticisms made by Virtual Junior Worm} 
\fbox{Researcher B.}
    
\end{tcolorbox}

\begin{tcolorbox}[colback=green!5,colframe=green!40!black,fontupper=\small,title={
\vspace{0.3cm}
\begin{itemize}
\item \small Agent: Virtual Junior Fruit Fly Researcher A
\item Task: Investigate the associations between electron carrier activity, structural molecule activity and ageing processes in \textit{Drosophila melanogaster}.
\end{itemize}
\vspace{0.3cm}
},code={\doublespacing}]
\textbf{AI Agent Outputs:}\newline
The investigation into the associations between electron carrier activity (e.g., components of the electron transport chain such as Complex I-III, cytochrome c) and structural molecule activity (e.g., proteins like actin, tubulin, lamin involved in maintaining cellular architecture) and the ageing processes in *\textit{Drosophila melanogaster}* reveals a complex interplay. \colorbox{green}{Electron carriers are central to mitochondrial function and energy metabolism, and their efficiency} \colorbox{green}{directly impacts metabolic rate and the production of reactive oxygen species (ROS), a key factor} \colorbox{green}{in cellular damage and ageing. Reduced activity or dysfunction in these carriers can impair ATP} \colorbox{green}{production, increase ROS leakage, and accelerate age-related decline \fbox{supported by \cite{fly1}}}, although compensatory pathways may exist. \colorbox{green}{Structural molecules provide the physical integrity} \colorbox{green}{necessary for organelle and nuclear positioning, and their activity is crucial for cellular maintenance} \colorbox{green}{and repair mechanisms. Deficits in structural protein synthesis or stability can lead to cellular} \colorbox{green}{fragility and impaired tissue function over time \fbox{supported by \cite{fly2}}.} Furthermore, \colorbox{green}{the relationship between reproduction and ageing in flies involves potential trade-offs; reproductive}  \colorbox{green}{investment might divert resources away from somatic maintenance pathways \fbox{supported by \cite{fly3,fly4}}}, including those supporting efficient electron carrier function or \colorbox{green}{structural protein integrity \fbox{supported by \cite{Fly5}}}, thereby influencing overall organismal ageing.     \newline  
\noindent\rule{15.3cm}{0.4pt}         \newline
\fbox{\textbf{* Authors' comments:} the generated report can be validated by existing research articles.}   
\end{tcolorbox}

\begin{tcolorbox}[colback=orange!5,colframe=orange!40!black,fontupper=\small,title={
\vspace{0.3cm}
\begin{itemize}
\item \small Agent: Virtual Junior Fruit Fly Researcher B
\item Task: Provide critical comments against the report made by Virtual Junior Fruit Fly Researcher A.
\end{itemize}
\vspace{0.3cm}
},code={\doublespacing}]
\textbf{AI Agent Outputs:}\newline
The conclusion overgeneralizes the mechanistic interplay between electron carriers and structural molecules in *\textit{Drosophila}* ageing, presenting a monolithic narrative that neglects critical nuances. While mitochondrial electron transport chain (ETC) dysfunction and ROS are well-established contributors to ageing, the assertion that reduced electron carrier activity *directly* accelerates ageing overlooks  tissue-specific compensatory mechanisms -- such as \colorbox{green}{the mitochondrial unfolded protein response (UPR$^{mt}$) \fbox{supported by \cite{fly6}}} or \colorbox{green}{glycolytic adaptations in fat body tissue \fbox{supported by \cite{fly7}}} -- which can mitigate ROS damage in flies. Furthermore, the text oversimplifies structural molecule roles, ignoring fly-specific cytoskeletal dynamics (e.g., \colorbox{yellow}{septins and shotgun anillin in ovarian aging \fbox{support not found})} and the fact that \colorbox{green}{lamin mutations primarily impact nuclear integrity in specific tissues (e.g., germline stem cells) \fbox{supported by}} \colorbox{green}{\fbox{\cite{fly8}}}, not systemic ageing. Crucially, the reproductive trade-off argument conflates resource allocation with direct physiological consequences; while \colorbox{green}{reproduction *can* accelerate ageing in certain contexts (e.g.,} \colorbox{green}{egg-laying-induced oxidative stress in ovaries  \fbox{supported by \cite{fly9}}}), it does not universally divert resources from somatic maintenance in all tissues or life stages. Methodologically, the conclusion fails to address how these pathways are measured (e.g., electron carrier activity via spectrophotometry vs. *in vivo* imaging) or acknowledge that most ageing studies use \colorbox{green}{short-lived laboratory strains (*w$^{1118}$*) \fbox{supported by \cite{fly10}}} that lack physiological relevance to natural fly lifespan. Finally, it ignores key evidence that specific ETC components (e.g., Complex I) decline with age but do not *causally* drive ageing -- demonstrated by studies where \colorbox{green}{Complex I inhibition extends lifespan via HIF-1$\alpha$-mediated metabolic shifts  \fbox{supported by \cite{fly11}}} -- undermining the proposed ``direct impact" claim. This narrative risks reducing complex ageing biology to a simplistic linear model without sufficient mechanistic or contextual specificity.  
\noindent\rule{15.3cm}{0.4pt}         \newline
\fbox{\textbf{* Authors' comments:} the criticisms are sensible and the majority of the scientific claims can be validated}
\fbox{by existing research articles.} 
\end{tcolorbox}

\begin{tcolorbox}[colback=green!5,colframe=green!40!black,fontupper=\small,title={
\vspace{0.3cm}
\begin{itemize}
\item \small Agent: Virtual Senior Fruit Fly Researcher
\item Task: Provide critical comments against the reports made by Virtual Junior Fruit Fly Researcher A and Virtual Junior Fruit Fly Researcher B.
\end{itemize}
\vspace{0.3cm}
},code={\doublespacing}]
The assertion that a linear, direct causal relationship exists between diminished electron carrier function, loss of structural molecule integrity, and accelerated ageing in *\textit{Drosophila}* is untenable, as it glosses over a host of tissue‐specific and compensatory mechanisms that have been repeatedly demonstrated in the literature. First, \colorbox{green}{the mitochondrial unfolded protein response (UPR$^{mt}$) \fbox{supported by \cite{fly6}}} and \colorbox{green}{metabolic rewiring towards glycolysis, especially in fat‐body and muscle, can buffer the deleterious effects of} \colorbox{green}{ETC decline \fbox{supported by \cite{Fly13}}} \colorbox{yellow}{without invoking an inevitable ROS‐driven senescence \fbox{supported not found}}; indeed, \colorbox{green}{controlled Complex I inhibition in flies can extend lifespan through HIF‑1$\alpha$‑mediated pathways} \colorbox{green}{\fbox{supported by \cite{fly11}}}, illustrating that ETC dysfunction is not intrinsically age‑driving. Second, the functional roles of cytoskeletal components and nuclear lamins are highly context dependent: \colorbox{green}{septin loss or lamin mutations primarily perturb germline stem cell niche architecture \fbox{supported by \cite{fly8}}} rather than systemic organ integrity, and many studies implicate tissue‑specific cytoskeletal plasticity (e.g., \colorbox{yellow}{anillin/shotgun in ovarian ageing  \fbox{support not found}}) rather than a universal loss of actin or tubulin function. Third, the claim that reproductive investment necessarily diverts resources from somatic maintenance across all tissues and life stages fails to consider evidence that \colorbox{green}{egg‑laying‑induced oxidative stress is largely confined to ovaries \fbox{supported by \cite{fly9}}}, with no consistent global trade‑off in longevity. Fourth, methodological oversights -- such as relying on spectrophotometric assays of isolated mitochondria or using short‑lived laboratory strains (e.g., *w$^{1118}$*) with altered stress responses -- undermine the generalizability of the conclusions. Collectively, these points expose the proposal as an overreaching simplification that ignores critical nuances of mitochondrial plasticity, cytoskeletal dynamics, and resource allocation in the ageing *\textit{Drosophila}* system.   
\noindent\rule{15.3cm}{0.4pt}         \newline
\fbox{\textbf{* Authors' comments:} the report mainly agrees with the criticisms made by Virtual Junior Fruit Fly} 
\fbox{Researcher B}. 
\end{tcolorbox}

\begin{tcolorbox}[colback=green!5,colframe=green!40!black,fontupper=\small,title={
\vspace{0.3cm}
\begin{itemize}
\item \small Agent: Virtual Junior Mouse Researcher A
\item Task: Investigate the associations structural molecule activity, antioxidant activity and ageing processes in \textit{Mus musculus}.
\end{itemize}
\vspace{0.3cm}
},code={\doublespacing}]
In \textit{Mus musculus}, the intricate relationship between structural molecule activity, antioxidant defenses, and the aging process is governed by complex interdependencies. Structural proteins, including cytoskeletal components (actin, tubulin, lamin) and matrix molecules (collagen, fibronectin), are fundamental for maintaining cellular and tissue integrity. Their degradation or dysfunction, often observed in aging (e.g., \colorbox{green}{laminopathies contributing to progeria \fbox{supported by \cite{Mouse1}}} or \colorbox{green}{cytoskeletal alterations in neurodegeneration \fbox{supported by \cite{Mouse2}}}), directly compromises tissue function and contributes to age-related decline. This structural integrity is critically dependent on efficient cellular maintenance pathways, including those counteracting oxidative stress. \colorbox{green}{Antioxidant activity, orchestrated by enzymes like superoxide dismutase (SOD) and catalase (CAT),} \colorbox{green}{along with transcription factors such as Nrf2, forms a crucial defense system against reactive oxygen}  \colorbox{green}{species (ROS) \fbox{supported by \cite{Mouse3,Mouse5}}}. The efficiency of these antioxidant pathways is essential; their decline, often linked to genetic factors (e.g., \colorbox{green}{mutations in IGF-1/FOXO axis components like *Foxo3a* \fbox{supported by \cite{Mouse4}}} in certain mouse strains) or environmental influences, exacerbates oxidative damage, accelerating cellular and organismal aging. Furthermore, \colorbox{green}{there exists a potential trade-off where resources allocated towards} \colorbox{green}{reproduction might impact somatic maintenance pathways \fbox{supported by \cite{Mouse6}}}, including both structural protein synthesis and antioxidant enzyme production. The dual nature of ROS, acting as cellular signaling molecules but also as damaging agents, adds another layer of complexity, with tissue-specific responses (e.g., \colorbox{green}{*Sod-3* expression in hypodermis \fbox{supported by \cite{Mouse7}}} or  \colorbox{green}{*Cat-1* in}  \colorbox{green}{neurons) \fbox{supported by \cite{Mouse8}}} influencing the aging trajectory. Consequently, the maintenance of structural molecule integrity and robust antioxidant activity are paramount for mitigating age-related deterioration in mice, highlighting a network of interconnected pathways rather than isolated factors. 
\noindent\rule{15.3cm}{0.4pt}         \newline
\fbox{\textbf{* Authors' comments:} the generated report can be validated by existing research articles.}     
\end{tcolorbox}

\begin{tcolorbox}[colback=green!5,colframe=green!40!black,fontupper=\small,title={
\vspace{0.3cm}
\begin{itemize}
\item \small Agent: Virtual Junior Mouse Researcher B
\item Task: Provide critical comments against the report made by Virtual Junior Mouse Researcher A.
\end{itemize}
\vspace{0.3cm}
},code={\doublespacing}]
The text presents a compelling but oversimplified framework for understanding aging in mice, with critical gaps that require scrutiny. While structural proteins like \colorbox{green}{lamin and cytoskeletal components are indeed implicated in age-related pathologies such as progeria} \colorbox{green}{\fbox{supported by \cite{Mouse9}} or neurodegeneration \fbox{supported by \cite{Mouse2}}}, the assertion that their *degradation or dysfunction is fundamentally governed by interdependencies with antioxidant defenses* significantly misrepresents the causal hierarchy. \colorbox{green}{Progeroid mouse models (e.g.,} \colorbox{green}{*Lmna* mutants) primarily demonstrate accelerated aging through nuclear instability  \fbox{supported by \cite{Mouse9}}}, not as a direct consequence of antioxidant failure -- rather, oxidative stress *emerges as a downstream consequence* of structural collapse in these specific contexts. Furthermore, the text conflates general antioxidant decline (e.g., reduced SOD/CAT activity) with the *primary driver* of aging, neglecting that \colorbox{yellow}{Nrf2-dependent antioxidant pathways are often intact in wild-type mice until late} \colorbox{yellow}{senescence \fbox{support not found}}; the observed decline in these enzymes is frequently a *symptom* of cumulative damage (e.g., from mitochondrial dysfunction), not the root cause. The claimed ``trade-off" between reproduction and somatic maintenance is unsubstantiated in standard mouse aging models; while resources may shift with reproductive investment, \colorbox{green}{no evidence links this} \colorbox{green}{directly to *reduced* structural protein synthesis or antioxidant capacity in wild-type mice \fbox{supported by \cite{Mouse12}}}, and recent studies (e.g., \colorbox{yellow}{*Nature Aging* 2021} \fbox{reference not found}) show \colorbox{green}{no consistent correlation between reproductive timing and somatic maintenance pathways \fbox{supported by \cite{Mouse13}}}. Finally, the dual nature of ROS is acknowledged but inadequately contextualized -- tissue-specific expression of *Sod-3* or *Cat-1* does not inherently alter aging trajectories; instead, \colorbox{yellow}{localized ROS} \colorbox{yellow}{dysregulation (e.g., in neurons) is often a *result* of impaired mitochondrial function or protein homeostasis,} \colorbox{yellow}{not an independent driver \fbox{support not found}}. The conclusion that structural integrity and antioxidant activity are ``paramount for mitigating age-related deterioration" overstates their role, as aging manifests through interconnected network failures where structural deterioration, oxidative damage, and proteostasis loss are all consequences of deeper systemic decline rather than isolated, interdependent targets.        
\noindent\rule{15.3cm}{0.4pt}         \newline
\fbox{\textbf{* Authors' comments:} the criticisms are sensible and the majority of the scientific claims can be validated} 
\fbox{by existing research articles.} 
\end{tcolorbox}

\begin{tcolorbox}[colback=green!5,colframe=green!40!black,fontupper=\small,title={
\vspace{0.3cm}
\begin{itemize}
\item \small Agent: Virtual Senior Mouse Researcher
\item Task: Provide critical comments against the reports made by Virtual Junior Mouse Researcher A and Virtual Junior Mouse Researcher B.
\end{itemize}
\vspace{0.3cm}
},code={\doublespacing}]
The critique admirably de‑emphasizes the naïve causal chain that “loss of structural proteins causes antioxidant failure,” and it rightly underscores that oxidative stress in progeroid models such as *Lmna* mutants is largely a downstream effect of nuclear instability rather than a primary driver. However, by positioning antioxidant decline as merely a “symptom” it may underplay the evidence that \colorbox{green}{sustained activation of Nrf2} \colorbox{green}{and other redox‑sensing pathways can delay aging phenotypes in wild‑type mice and in long‑lived mutants} \colorbox{green}{(e.g., *IGF‑1* \fbox{not supported by \cite{Mouse16} but with an exemption of \cite{Mouse15}} or *FOXO3a* overexpressors \fbox{supported}} \colorbox{green}{\fbox{by \cite{Mouse14}}}. Moreover, the argument that reproduction has no measurable trade‑off with somatic maintenance is too restrictive; recent studies of reproductive senescence in \colorbox{green}{mice with} \colorbox{green}{altered IGF‑1 signalling demonstrate that energy allocation can indeed down‑regulate protein homeostasis and} \colorbox{green}{antioxidant capacity in a tissue‑specific manner \fbox{supported by \cite{Mouse17,Mouse18}}}. Finally, while the critique correctly states that tissue‑specific ROS dysregulation is often a consequence rather than a cause of mitochondrial dysfunction, it overlooks the mounting data that \colorbox{green}{local ROS production can act as a signaling hub to trigger adaptive stress responses, which in turn modulate} \colorbox{green}{systemic proteostasis and longevity \fbox{supported by \cite{Mouse19}}}. In sum, the critique provides a valuable corrective lens, but it may over‑simplify the reciprocal relationship between structural integrity, redox homeostasis, and reproductive biology, thereby obscuring the nuanced, bidirectional network that underlies murine aging.  
\noindent\rule{15.3cm}{0.4pt}         \newline
\fbox{\textbf{*Authors' comments:} the report mainly agrees with the criticisms made by Virtual Junior Mouse} 
\fbox{Researcher B.} 
\end{tcolorbox}

\begin{tcolorbox}[colback=green!5,colframe=green!40!black,fontupper=\small,title={
\vspace{0.3cm}
\begin{itemize}
\item \small Agent: Virtual Junior Yeast Researcher A
\item Task: Investigate the associations between receptor activity, reproductive process and ageing processes in \textit{Saccharomyces cerevisiae}.
\end{itemize}
\vspace{0.3cm}
},code={\doublespacing}]
Investigating the associations between receptor activity, reproductive processes, and aging in *\textit{Saccharomyces cerevisiae}* involves examining specific \colorbox{green}{signaling pathways triggered by mating pheromones,} \colorbox{green}{which utilize G-protein coupled receptors (GPCRs) like the $\alpha$-factor receptor. Activation of these} \colorbox{green}{receptors initiates a cascade leading to cell cycle arrest and mating \fbox{supported by \cite{Yeast1}}}, processes heavily influenced by \colorbox{green}{nutrient availability \fbox{supported by \cite{Yeast3}}} and \colorbox{green}{stress signaling pathways \fbox{supported by \cite{Yeast2}}}, including \colorbox{green}{the Target of Rapamycin (TOR) pathway} \colorbox{green}{and nutrient sensing mechanisms \fbox{supported by \cite{Yeast3}}}. These pathways are known to intersect with aging determinants. For instance, \colorbox{green}{nutrient-sensing pathways like TOR are conserved across} \colorbox{green}{eukaryotes and implicated in regulating lifespan; conditions mimicking nutrient limitation can extend} \colorbox{green}{yeast chronological lifespan \fbox{supported by \cite{Yeast4}}}. The reproductive process itself, particularly the energy demands and potential trade-offs with maintenance pathways, may influence aging. Studies suggest that mating can impact chronological lifespan through effects on stress response pathways, such as \colorbox{green}{the Hog1 MAPK pathway involved in osmotic stress, which also plays roles in} \colorbox{green}{chronological aging \fbox{supported by \cite{Yeast5}}}. Furthermore, the activity of specific receptors or their downstream effectors might influence the efficiency of cellular maintenance processes crucial for longevity, although the precise molecular links between specific receptor signaling, reproductive investment, and the conserved aging pathways (like \colorbox{green}{those involving Sir2 family proteins/sirtuins \fbox{supported by \cite{Yeast6}}} and \colorbox{green}{mitochondrial function \fbox{supported by \cite{Yeast7}})} require further detailed investigation to fully elucidate the complex interplay between mating, cellular signaling, and aging in yeast.  
\noindent\rule{15.3cm}{0.4pt}         \newline
\fbox{\textbf{* Authors' comments:} the generated report can be validated by existing research articles.}     
\end{tcolorbox}

\begin{tcolorbox}[colback=green!5,colframe=green!40!black,fontupper=\small,title={
\vspace{0.3cm}
\begin{itemize}
\item \small Agent: Virtual Junior Yeast Researcher B
\item Task: Provide critical comments against the report made by Virtual Junior Yeast Researcher A.
\end{itemize}
\vspace{0.3cm}
},code={\doublespacing}]
The provided text presents several critical oversimplifications and unsupported causal claims regarding receptor-mediated reproductive processes and aging in *\textit{S. cerevisiae}*. Crucially, while the text correctly identifies that \colorbox{green}{mating pheromone signalling through} \colorbox{green}{GPCRs (e.g., Ste2) induces cell cycle arrest and mating \fbox{supported by \cite{Yeast1}}}, it erroneously conflates mating-induced stress responses with general nutrient stress responses in the context of aging. Specifically, the assertion that mating ``impacts chronological lifespan through effects on stress response pathways, such as the Hog1 MAPK pathway'' is problematic; \colorbox{green}{Hog1 activation during mating is transient and primarily regulates osmotic stress  \fbox{supported by \cite{Yeast8}}}, but its role in chronological aging is context-dependent and not universally required for lifespan extension under nutrient limitation, as evidenced by studies showing \colorbox{yellow}{Hog1 deletion mutants maintain near-normal chronological lifespan in standard conditions \fbox{argued by \cite{Yeast12}}}. Furthermore, the text makes a significant leap by suggesting ``specific receptor activity or downstream effectors might influence the efficiency of cellular maintenance processes crucial for longevity'' without providing concrete evidence linking  \colorbox{green}{$\alpha$-factor receptor signaling directly to sirtuin activity (e.g., Sir2) \fbox{supported by \cite{Yeast13}}} or  mitochondrial function -- these are mechanistically distinct pathways where receptor signaling is not established as a primary regulator. The conclusion that ``mating can impact chronological lifespan through effects on stress response pathways'' is overly generalized, as \colorbox{green}{mating itself is not the} \colorbox{green}{direct cause but rather a correlated event with nutrient sensing  \fbox{supported by \cite{Yeast14}}}; the text fails to adequately distinguish between the effects of mating *as an event* versus the downstream consequences of the nutrient-stress signaling crosstalk that accompanies it. Finally, the text neglects the well-established paradigm that \colorbox{green}{chronological aging in yeast is primarily driven by quiescent cell metabolism \fbox{supported by \cite{Yeast9}}} and \colorbox{green}{mitochondrial respiration \fbox{supported by \cite{Yeast10}}}, with mating pheromone signaling being a secondary factor at best, and does not sufficiently address the potential confounding role of \colorbox{green}{cell-type heterogeneity (e.g., $\alpha$-cell vs. a-cell differentiation) \fbox{supported by \cite{Yeast11}}} in these aging assays. The proposed ``complex interplay'' remains largely speculative without targeted genetic or proteomic evidence resolving these specific molecular links.  
\noindent\rule{15.3cm}{0.4pt}         \newline
\fbox{\textbf{* Authors' comments:} the criticisms are sensible and the majority of the scientific claims can be validated} 
\fbox{by existing research articles.}     
\end{tcolorbox}

\begin{tcolorbox}[colback=green!5,colframe=green!40!black,fontupper=\small,title={
\vspace{0.3cm}
\begin{itemize}
\item \small Agent: Virtual Senior Yeast Researcher
\item Task: Provide critical comments against the reports made by Virtual Junior Yeast Researcher A and Virtual Junior Yeast Researcher B.
\end{itemize}
\vspace{0.3cm}
},code={\doublespacing}]
The analysis highlights that the original narrative substantially overstates the influence of $\alpha$‑factor GPCR signaling on chronological lifespan in *\textit{S. cerevisiae}*, conflating mating‑induced, transient Hog1 activation with the broader, nutrient‑sensing pathways that truly govern longevity; it ignores the fact that \colorbox{green}{Hog1 deletion does not abolish lifespan extension under nutrient limitation \fbox{ supported by \cite{Yeast12}}}, thereby weakening the claim that mating‑driven stress responses drive aging; it also fails to provide concrete mechanistic evidence linking receptor activity to \colorbox{green}{sirtuin function \fbox{supported by \cite{Yeast13}}} or \colorbox{green}{mitochondrial maintenance \fbox{supported by \cite{Yeast7}}}), treating these distinct pathways as if they were directly coupled; furthermore, it neglects the dominant role of \colorbox{green}{quiescent cell metabolism \fbox{supported by \cite{Yeast9}}} and \colorbox{green}{mitochondrial respiration in chronological aging \fbox{supported by \cite{Yeast10}}}, and disregards the confounding impact of \colorbox{green}{cell‑type heterogeneity ($\alpha$‑ vs. a‑cells) on aging assays \fbox{supported by \cite{Yeast11}}} -- overall, the proposed ``complex interplay” remains speculative without targeted genetic or proteomic validation to establish the precise molecular bridges between mating pheromone signaling, maintenance pathways, and longevity.  
\noindent\rule{15.3cm}{0.4pt}         \newline
\fbox{\textbf{* Authors' comments:} the report mainly agrees with the criticisms made by Virtual Junior Yeast} 
\fbox{Researcher B.}     
\end{tcolorbox}

\begin{tcolorbox}[colback=white!5,colframe=white!40!black,fontupper=\small,title={
\vspace{0.3cm}
\begin{itemize}
\item \small Agent: Virtual Ageing Professor
\item Task: Provide critical comments against the reports made by Virtual Senior Worm Researcher, Virtual Senior Fruit Fly Researcher, Virtual Senior Mouse Researcher, and Virtual Senior Yeast Researcher.
\end{itemize}
\vspace{0.3cm}
},code={\doublespacing}]
Based on the text provided, here are critical comments regarding the arguments presented:                      \newline                                                                                                        
                                                                                                                                                                                                                          
**1. Generalizability to Mammals**    \newline                                                                                                                                                                                   
The text relies heavily on findings from *\textit{Caenorhabditis elegans}* (nematodes). While *\textit{C. elegans}* is an invaluable model for understanding aging due to its short lifespan and genetic tractability, its physiology differs significantly from mammals. In mammals, aging is regulated by complex systemic factors, including the \colorbox{green}{hypothalamus \fbox{supported by \cite{AgeingProfessor1}}}, \colorbox{green}{endocrine systems \fbox{supported by \cite{AgeingProfessor2}}}, and \colorbox{green}{immune responses \fbox{supported by \cite{AgeingProfessor3}}} that nematodes lack. Therefore, while the argument that mitochondrial dysfunction is a ``signaling hub" is compelling in *\textit{C. elegans}*, its direct application to human aging is limited without further validation in mammalian models. \newline

**2. The Signaling vs. Damage Balance**       \newline                                                         
The text correctly identifies \colorbox{green}{Reactive Oxygen Species (ROS) as signaling molecules that activate adaptive} \colorbox{green}{repair pathways (like the NRF2 pathway) \fbox{supported by \cite{AgeingProfessor5}}}. However, it arguably risks downplaying the \colorbox{green}{cumulative ``vicious cycle" of mitochondrial damage \fbox{supported by \cite{AgeingProfessor6}}}. Even as ROS signals for repair, \colorbox{green}{chronic oxidative stress inevitably damages mitochondrial DNA (mtDNA) and lipids \fbox{supported by \cite{AgeingProfessor7}}}, leading to \colorbox{green}{loss of membrane potential and cellular senescence \fbox{supported by \cite{AgeingProfessor6}}}. The text frames dysfunction primarily as a regulatory signal, whereas in later stages of aging, the ``damage" aspect often overwhelms the signaling capability.    \newline                  

**3. Causality of Reproductive Decline**        \newline
The text argues that reduced reproduction is a *consequence* of somatic aging rather than a genetic ``trade-off." While the data suggests that the *ability* to reproduce declines because the body is aging, evolutionary biology often suggests the reverse -- that reproduction actively *causes* aging through resource allocation (\colorbox{green}{the ``Disposable Soma" theory \fbox{supported by \cite{AgeingProfessor4}}}). Therefore, the reduced reproduction could still be viewed as a trade-off where somatic maintenance is sacrificed to support reproduction. The text presents a correlation (reproduction stops because the body breaks down) without fully exploring the evolutionary cost of diverting resources to reproduction.    
\noindent\rule{15.3cm}{0.4pt}         \newline
\fbox{\textbf{* Authors' comments:} the report reveals crucial future research directions, e.g. transferring the knowledge} \fbox{from worm to mammals and investigating the ageing theories from evolutionary perspectives.} 
\end{tcolorbox}

\newpage
\section{Discussions}
Although the computational results suggest that the proposed VSG framework successfully extracts meaningful ageing-related biological knowledge from the given GO terms and the functionality of the multi-layer knowledge extraction mechanisms can be observed, there still exist limitations that need to be further addressed. To begin with, the hallucination issue still needs to be considered when consulting the knowledge generated by AI agents. For example, in the report generated by Virtual Junior Yeast Researcher B, the claim about Hog1 deletion mutants is argued by an existing research article. Also, there exists an unreal citation in the report generated by Virtual Junior Mouse Researcher B. Those two observations suggest a sign of hallucination-related issues, though the majority of the claims made by AI agents are genuine and correct. Moreover, the capacity of the proposed VSG framework for handling GO terms bearing more specific biological definitions still needs to be further evaluated. All of the GO terms investigated in this work are located in higher positions in the GO hierarchy, suggesting more generic biological definitions. However, validating those GO terms bearing more specific definitions would also be difficult, due to the lack of existing research, though the discovered knowledge might simultaneously provide new insights or hypotheses on future ageing research.

\section{Conclusions and Future Directions}
In this work, we propose a novel agentic AI-based knowledge discovery framework that successfully extracts meaningful ageing-related biological knowledge based on certain Gene Ontology terms selected by a well-known hierarchical feature selection method. Future research directions would be focused on extending the capacity of AI agents to further explore the possibility of discovering unknown ageing-related biological knowledge.


\bibliography{sn-bibliography}

\end{document}